\documentclass[letterpaper, 10 pt, conference]{ieeeconf}  

\IEEEoverridecommandlockouts                           
\usepackage{graphicx} 
\usepackage{mathptmx} 
\usepackage{times} 
\usepackage{amsmath} 
\usepackage{amssymb}  
\usepackage{xcolor} 
\usepackage{xurl} 
\usepackage{caption}

\usepackage{subcaption}
\usepackage{booktabs}
\makeatletter
\let\NAT@parse\undefined
\makeatother
\usepackage{hyperref}

\title{\LARGE \bf
Enhancing Visual Perception in Novel Environments via Incremental Data Augmentation Based on Style Transfer
}

\author{Abhibha Gupta$^{1}$, Rully Agus Hendrawan$^{2}$, Mansur Arief$^{3}$
\thanks{*This research received no specific grant from any funding agency}
\thanks{$^{1}$Abhibha Gupta is with the University of Pittsburgh, Pittsburgh, PA 15213, USA {\tt\small abg96@pitt.edu}}%
\thanks{$^{2}$Rully Agus Hendrawan is with the University of Pittsburgh, Pittsburgh, PA 15213, USA and Institut Teknologi Sepuluh Nopember, Sukolilo, Surabaya 60111, Indonesia {\tt\small ruhendrawan@gmail.com}}%
\thanks{$^{3}$Mansur Arief is with the Stanford University, Stanford, CA 94305, USA {\tt\small ariefm@stanford.edu}}%
}

\begin{document}

\maketitle
\thispagestyle{empty}
\pagestyle{empty}

\begin{abstract}
The deployment of autonomous agents in real-world scenarios is challenged by "unknown unknowns", i.e. novel unexpected environments not encountered during training, such as degraded signs. While existing research focuses on anomaly detection and class imbalance, it often fails to address truly novel scenarios. Our approach enhances visual perception by leveraging the Variational Prototyping Encoder (VPE) to adeptly identify and handle novel inputs, then incrementally augmenting data using neural style transfer to enrich underrepresented data. By comparing models trained solely on original datasets with those trained on a combination of original and augmented datasets, we observed a notable improvement in the performance of the latter. This underscores the critical role of data augmentation in enhancing model robustness. Our findings suggest the potential benefits of incorporating generative models for domain-specific augmentation strategies.
\end{abstract}

\section{Introduction}

The deployment of autonomous robots on real-world tasks is often hindered by the lack of robustness to unexpected situations. For example, autonomous vehicles are expected to operate safely in complex environments, including in the presence of pedestrians, cyclists, and other vehicles. However, the real world is full of novel environments that were not anticipated during training, i.e., unknown unknowns. 

\begin{figure}
    \centering
    \includegraphics[width=0.6\linewidth]{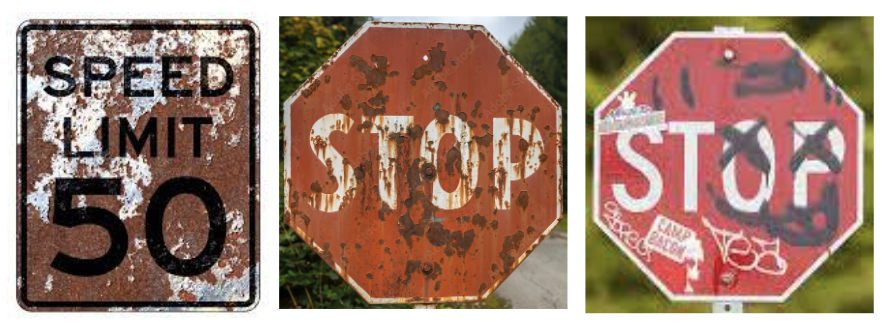}
    \caption{Examples of degraded traffic signs in the real-world}\label{fig:rusty_signs}
\end{figure}

Figure~\ref{fig:rusty_signs} shows examples of traffic signs in the real world that exhibit degradation often not seen in training datasets~\cite{houben2013detection}, such as fading colors or being invaded by rust or graffiti~\cite{eykholt2018robust}. Such degradation can significantly affect sign retroreflectivity and readability, even for humans~\cite{veneziano2023sign}. This unexpected issue, which is prevalent in practice due to the limited budget and complexity of traffic sign maintenance operations~\cite{chen2009assessment}, complicates the design of reliable perception systems~\cite{pei2017towards}. This issue is also faced by other autonomous agents that rely on computer vision, such as drones~\cite{lindqvist2021reactive} and robots~\cite{paxton2019visual}. Despite the risks posed by such unknown unknowns, limited attention has been paid to this class of problems in the literature~\cite{ramanishka2018toward}.

Although the type of unknown unknowns mentioned previously share some similarities with the well-studied problem of class imbalance~\cite{japkowicz2002class, he2009learning}, there are some key differences. Class imbalance occurs when training data for one class is scarce compared to others~\cite{buda2018systematic, van2007experimental}. However, at least some examples of the underrepresented class are available in the training set. In contrast, unknowns emerge when training data are not available for certain cases in the real world~\cite{sculley2014machine}. For instance, a traffic sign that has been heavily invaded by rust may not be present in the training set, and such cases will eventually occur during deployment. Arguably, the latter problem class is more challenging to address compared to the former problem class, as the model may have no prior knowledge, especially if the trained model does not have inherent out-of-distribution generalization or one-shot learning capabilities~\cite{vinyals2016matching, kim2019variational}.

Most of the work on this problem uses outlier or anomaly detection techniques~\cite{chalapathy2019deep, zhou2017anomaly}. Methods rely on reconstructing input and measuring reconstruction errors to find outliers~\cite{an2015variational}. While these approaches can detect unusual inputs, they do not provide information on the specific unknown situation. Other techniques learn a classifier boundary around known training data and detect outliers outside this boundary~\cite{ruff2018deep}. However, this approach is limited when unknown examples are near the class boundary. 

\begin{figure*}
    \centering
    \begin{subfigure}[b]{0.49\linewidth}
        \centering
        \includegraphics[width=\linewidth]{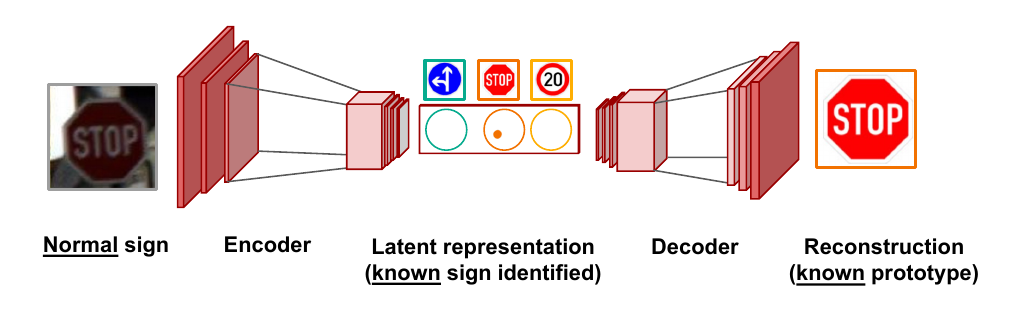}
        \caption{Normal traffic sign}\label{fig:success_reconstruction}
    \end{subfigure}
    \begin{subfigure}[b]{0.49\linewidth}
        \centering
        \includegraphics[width=\linewidth]{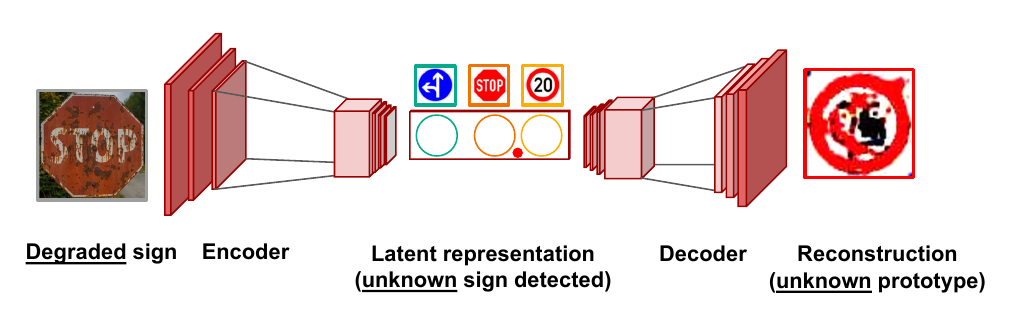}
        \caption{Degraded traffic sign}\label{fig:failed_reconstruction}
    \end{subfigure}
    \caption{VPE reconstruction examples for normal and degraded traffic sign}\label{fig:vpe_reconstructions}
\end{figure*}

In this study, we focus on a data augmentation approach exposing models to unfamiliar cases~\cite{shorten2019survey, hendrycks2019benchmarking}. Generative adversarial networks can automate the augmentation process~\cite{antoniou2017data}, but still rely on existing data distributions. Truly novel cases are difficult to simulate because generating useful new examples to retrain the model requires creativity and domain expertise. A promising one-shot detection based on a variational prototyping encoder (VPE)~\cite{kim2019variational} can detect novel or ambiguous inputs by mapping the input images into a latent representation in which images of the same class are mapped close together. However, this approach by itself does not provide a mechanism to adapt to unknown unknowns. Even lacking such capability, VPE can still be used as an effective novel or ambiguous class detector. Fig.~\ref{fig:vpe_reconstructions} shows examples of both VPE succeeding in classifying a normal stop sign (and can even reconstruct its prototype, i.e., its image symbol, Fig.~\ref{fig:success_reconstruction}) and another in which it detects a novel sign for a severely rusty stop sign (Fig.~\ref{fig:failed_reconstruction}, which yields a failed prototype reconstruction). 

To address this limitation, we propose an incremental data augmentation that combines VPE with neural style transfer (NST)~\cite{luan2017deep} and systematic data augmentation procedure~\cite{wang2016training, cubuk2019autoaugment}. At runtime, when VPE detects novel or ambiguous inputs, the autonomous agent shall warn the operator, flag the input, and optionally resort to a failsafe mechanism. The augmentation and retraining module is then kicked in, exposing the VPE model to novel environments by transferring the novel styles from these inputs to prototypes and the original training set via NST. The augmented prototypes are then used to retrain the VPE model efficiently~\cite{wang2016training}, enhancing its robustness to this specific degradation instance. The agent can then take over again when it finishes the retraining step and can correctly identify the degraded style. This loop of monitoring, augmenting, and adapting incrementally is expected to improve the robustness of the agent against unknown sign degradations. Compared to generic augmentation~\cite{cubuk2019autoaugment}, this approach is much more efficient since we tailor the data to the specific unknown styles encountered, therefore reducing the computational overhead to retrain the agent.

The contribution of this work is twofold. First, we propose a framework to address unknown unknowns in the real world by combining VPE, NST, and incremental augmentation. Second, we demonstrate the effectiveness of our approach to improve the robustness of visual perceptions in the context of degraded traffic sign classifications. We show that our approach can improve the robustness of the classifier against novel environments, paving the way toward building more reliable autonomous agents. The code is available at {\small\url{https://github.com/abhibha1807/Robustifying_Visual_Perception/}}.

\section{Related Work}
We briefly review the literature on robust visual perception models and data augmentation approaches.  We then discuss NST as a promising technique to generate realistic degradations for augmentation.

\subsection{Robust Visual Perception Models}
Visual perception is a critical component for autonomous agents that rely on sensors to safely navigate the environments, e.g. autonomous vehicles or advanced driver assistance systems. However, building robust classifiers that can handle complex real-world conditions remains an open challenge. In the wild, novel environments often emerge, even in semi-regularized settings, such as roadways with traffic signs. Traffic signs undergo degradations and perturbations often not present in clean datasets. As~\cite{varshney2016engineering} discuss, managing such unanticipated cases is key to reliable perception systems. This review of the literature summarizes recent progress in improving the robustness of traffic sign classifiers, focusing on approaches that aim to improve generalization and handle unknown or unexpected inputs at test time.

A seminal work by~\cite{ciregan2012multi} showed how deep neural networks could surpass human performance on standardized traffic sign datasets. However, unknown unknowns that emerge in the real world can still cause failures. Strategies to improve robustness include: (1) novel architectures such as variational prototyping encoder (VPE)~\cite{kim2019variational}, (2) adversarial training~\cite{eykholt2018robust} with simulated perturbations, and (3) data augmentation~\cite{hendrycks2019benchmarking} with synthetic corruptions. To this end, VPE networks reconstruct prototypes from real images as a pretraining step, and at test time, prototypes of novel signs can be accurately classified based on the pre-trained encoder. The intuition is that learning to normalize against real-world perturbations provides embeddings invariant to those variations. Meanwhile, adversarial training augments datasets with perturbed examples, generating, e.g., physical stickers to simulate graffiti and occlusion attacks. Data augmentation exposes models to unseen data variations. Model retraining using these synthetic corruptions can therefore improve robustness to unseen degradations. However, manually designing effective augmentations requires creativity and domain expertise.

Recent work has looked at meta-learning models that can rapidly adapt to new classes with few examples~\cite{vinyals2016matching}. This enables detecting and learning about unknown cases. However, meta-learning models still require some examples of the unknown class. Handling zero-shot unknown unknowns is still an open problem, highlighting that existing techniques have limitations in identifying unexpected situations in the wild. Advances in few-shot, open-set, and generalized learning may provide pathways toward more robust perception systems, but such an approach is yet to be explored.

\subsection{Data Augmentation Approaches for Visual Perception}

Traffic signs in natural environments exhibit significant degradation not present in standard datasets. Strategies like data augmentation aim to expose models to realistic variation during training. A recent survey about data augmentation techniques is available in~\cite{lewy2023overview}. However, generating plausible synthetic degradations requires creativity and domain expertise. Neural style transfer is a promising technique that could generate realistic corruptions by transferring textures from real degradation examples.

NST leverages deep convolutional networks to render novel images combining the content of one image with the style of another~\cite{gatys2016image}. Further methods enabled fast approximation~\cite{johnson2016perceptual} and control over spatial style variations~\cite{li2018closed}. Recent work has adapted style transfer to synthetic data generation tasks.~\cite{geirhos2018imagenet} stylized natural images to match the textures of artwork and applied this as data augmentation to improve model robustness.~\cite{mogelmose2014traffic} modeled physical dirt and wear accumulation for augmented sign training. While promising, directly modeling specific corruptions may miss unexpected degradation modes in the wild. Neural style transfer could expose models to diverse variations by transferring texture and damage patterns from real degraded signs.

\section{Methodology}

\begin{figure*}
    \centering
    \includegraphics[width=\linewidth]{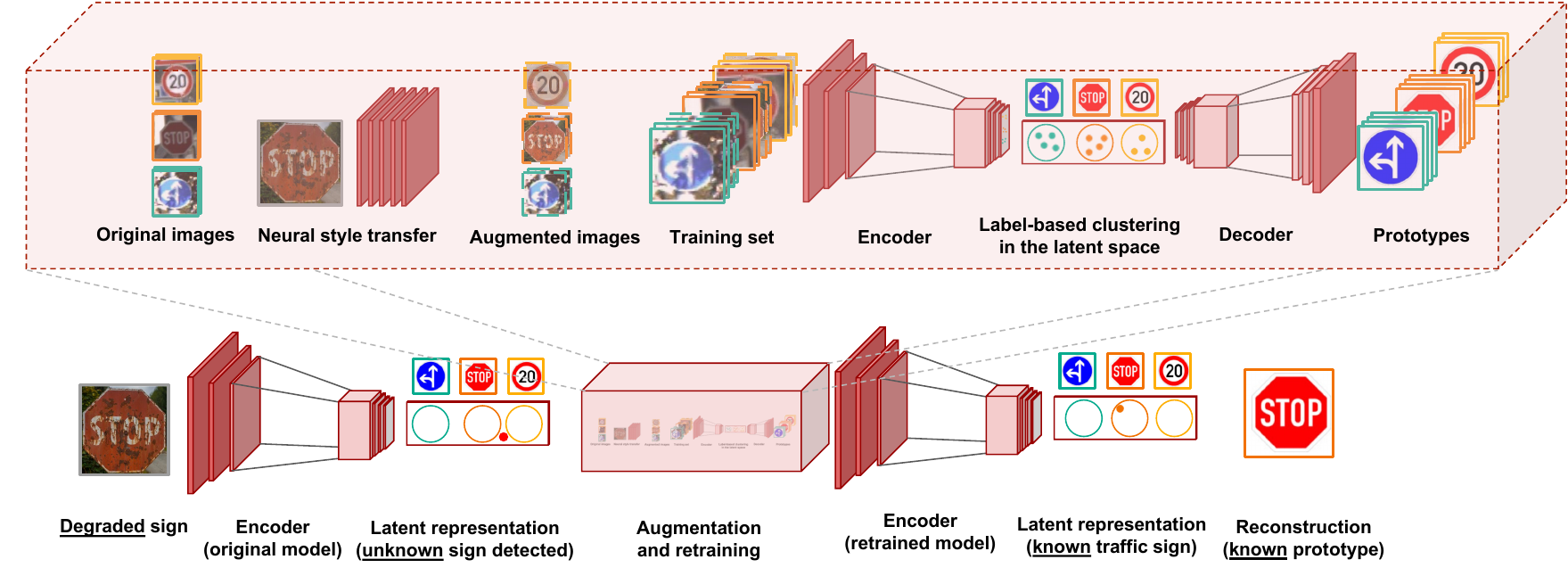}
    \caption{Illustration of the proposed framework}\label{fig:illustration}
\end{figure*}

To address the issue of unknown unknowns in visual perception, we propose an incremental augmentation framework based on  VPE~\cite{kim2019variational}, NST~\cite{luan2017deep}, and model retraining~\cite{wang2016training, cubuk2019autoaugment}. This framework is illustrated in Fig.~\ref{fig:illustration}

\subsection{Variational Prototype Encoding (VPE)}

At runtime, the VPE module serves as a monitor to detect unknown or ambiguous input. For each input image $\mathbf{x} \in \mathbb{R}^{H \times W \times 3}$, the VPE encodes it into a latent vector $\mathbf{z} \in \mathbb{R}^d$ aligned with prototype clusters
\begin{align}
\mathbf{z} = \text{VPE}(\mathbf{x}; \theta_{\text{VPE}}),
\end{align}
where $\theta_\text{VPE}$ denotes the parameters of the VPE encoder that we train using an original training set $\mathcal D$. The VPE is implemented as a convolutional neural network trained to map real-world traffic sign images to their ideal prototype representations. To measure the novelty, we use the distance $d$ between the nearest prototype cluster centroid $\mathbf{z}_{\text{proto}}$ and $\mathbf{z}$
\begin{align}
d &= \|\mathbf{z} - \mathbf{z}_{\text{proto}}\|_2
\end{align}
where in this case, the distance is measured in the VPE latent space. One can obtain the label of the closest prototype images from precompiled tabular data or via a simple linear classifier
\begin{align}
\hat{\mathbf{y}} &= \text{Softmax}(\mathbf{W}\mathbf{z}_{\text{proto}} + \mathbf{b}),
\end{align}
where $\mathbf W$ and $\mathbf b$ are the associated weights that can be appended into the VPE networks to obtain the final prediction. Images with $d > \tau_d$ or $\hat{y} = \max(\hat{\mathbf{y}}) < \tau_y$ are flagged as unfamiliar and stored in a buffer $\mathcal{B}$ for augmentation. The thresholds $\tau_d$, $\tau_y$ are selected by cross-validation.

\subsection{Neural Style Transfer (NST)}

The style transfer module transfers texture and degradation styles from unknown images collected $\mathbf{x}_{\text{unknown}} \in \mathcal{B}$ to prototypes $\mathbf{x}_{\text{proto}}$ to generate realistic augmented data $\mathbf{x}_{\text{aug}}$. For our traffic sign example, the rusty effects, faded colors, or graffiti styles of unfamiliar stop signs would be transferred to pristine stop sign prototypes:
\begin{align}
\mathbf{x}_{\text{aug}} = \text{StyleTransfer}(\mathbf{x}_{\text{unknown}}, \mathbf{x}_{\text{proto}}; \theta_{\text{ST}}, \xi),
\end{align}
where $\theta_{\text{ST}}$ denotes the parameters of the style transfer network, and $\xi$ is the random seed that controls the randomness of the generated image. Style transfer is built on top of DALL-E~\cite{marcus2022very}, which appears to disentangle content and style representations.

With this, we can sample $\xi$'s and use all samples in the buffer $\mathcal B$ to construct an augmentation set 
\begin{align}
\mathcal B^\prime = \{\mathbf{x}_{\text{aug}, i}, \hat{y}_i  \}_{i=1}^{n^\prime}, 
\end{align}
where $n^\prime$ is the number of tuples in the augmented data, exposing the model to unknown degradations. The final augmented dataset $\mathcal D^\prime$ is constructed by ensuring that $p \%$ samples are obtained from $\mathcal D$ and $1-p\%$ samples from $\mathcal B^\prime$. 

\subsection{Model Retraining}

The augmented dataset $\mathcal D^\prime$ is used to fine-tune the classifier $M_\theta$. For simplicity, we consider the VPE classifier as our classifier of choice, i.e., $\theta = \theta_{\text{VPE}}$. Hence, the fine-tuned model $M_{\theta^\prime}$ is obtained as
\begin{align}
M_{\theta^{'}} = \text{Retrain}(M_\theta, \mathcal D^\prime)
\end{align}
To this end, retraining $M_\theta$ with the new stylized data reduces unknown unknowns in future deployments. We iterate this process continuously - updating our model  $M_\theta \leftarrow M_\theta^\prime$, monitoring with VPE, stylizing prototypes, and retraining - to systematically address unknown unknowns and enhance the robustness of our traffic sign classifier.

The classifier $M_\theta$ can be implemented as any convolutional neural network parameterized by $\theta$. Fine-tuning is performed via stochastic gradient descent on the loss
\begin{align}
\mathcal{L}(\theta; \mathcal D, \mathcal D^\prime) = \mathcal{L}_{\text{CE}}(\theta; \mathcal D^\prime) + \lambda \mathcal{L}_{\text{Consist}}(\theta; \mathcal D),
\end{align}
where $\mathcal{L}_{\text{CE}}$ is the cross-entropy loss that encourages accurate predictions over the augmented dataset, $\mathcal{L}_{\text{Consist}}$ is the inconsistency loss that encourages consistency between predictions on original images, to prevent catastrophic forgetting~\cite{kirkpatrick2017overcoming}, and $\lambda$ is the weight.

\subsection{Implementation Details}
We implement our framework in PyTorch. The VPE network is based on~\cite{kim2019variational}. The style transfer network follows~\cite{geirhos2018imagenet}. For computational efficiency, we let the VPE classifier be our classifier of choice, so the retraining process updates both the VPE encoder and classifier simultaneously. This also implies that $\mathcal{L}_{\text{CE}}$ incorporates all VPE losses that include reconstruction error and distribution regularization term~\cite{kim2019variational}. During deployment, the VPE encoder and its classifier modules are required to monitor novel degraded signs and to perform robust predictions. Hyperparameters like loss weights and training schedules are tuned via random search. We evaluate our approach to GTSRB traffic sign datasets~\cite{houben2013detection}, corrupted with real-world degradations. Ablation studies analyze the contribution of each component. We demonstrate that directly adapting to unknown unknowns at runtime leads to more robust performance than the persistent model trained using the original training set alone.

\section{Experiments}

We experimentally evaluate our proposed framework on traffic sign classification using the German Traffic Sign Recognition Benchmark (GTSRB) dataset~\cite{houben2013detection}.

\subsection{Dataset}
GTSRB contains over 50,000 images across 43 classes of traffic signs. We generate a corrupted test set by applying degradation due to rust from real-world images using NST. This emulates the unknown unknowns encountered at deployment.
We introduce a novel dataset designed for evaluating image classification methods under different conditions of original and augmented imagery. The dataset consists of a total of 39,214 'Original' images, which are unmodified images, and 39,214 style transfer augmented (ST-Aug) or 'Rustic' images, obtained through the application of NST technique on the original dataset. Furthermore, the test set consists of 12630 original images, along with 12630 rustic test images generated using a similar NST approach.

\subsection{Scenarios}
We present an evaluation set of our proposed method's ability to address unknown unknowns and dataset imbalance in real-life situations. Our investigation involves training and testing on various combinations of original and augmented datasets, aiming to simulate diverse scenarios encountered in practical applications. We conduct six distinct experiments to establish baselines and assess the efficacy of our approach. 

\begin{enumerate}
\item Experiment 1: As our primary baseline, both training and testing exclusively use original images. It provides a reference point for optimal performance when there is no data augmentation, degradation, or other external challenges. By understanding how models perform under familiar conditions, we can measure the impact of these challenges in other experiments.

\item Experiment 2: This acts as our secondary baseline. Training occurs with original images, while testing uses augmented images. This reveals the challenges faced by current models trained on unperturbed datasets when encountering unknown-unknown cases.

\item Experiment 3: We evaluate the model's performance when trained on a combination of original and augmented data, then tested on original images. This model is trained on an environment that mimics a real-world scenario that has both augmented (e.g., rustic) and unaltered images, and tested on the common cases of original images.

\item Experiment 4: Training uses both original and augmented data, but is tested on the augmented set. This assesses the robustness of our technique in detecting images accurately, i.e., handling unknown unknowns.

\item Experiment 5: The model is trained solely on augmented images and tested on original images, examining the viability of training with only augmented data.

\item Experiment 6: Both training and testing utilize augmented data, simulating a worst-case scenario with limited quality datasets. In certain environments or applications, the majority of data encountered may be degraded or differ from standard datasets.

\end{enumerate}

\begin{figure}
    \centering
    \includegraphics[width=0.8\linewidth]{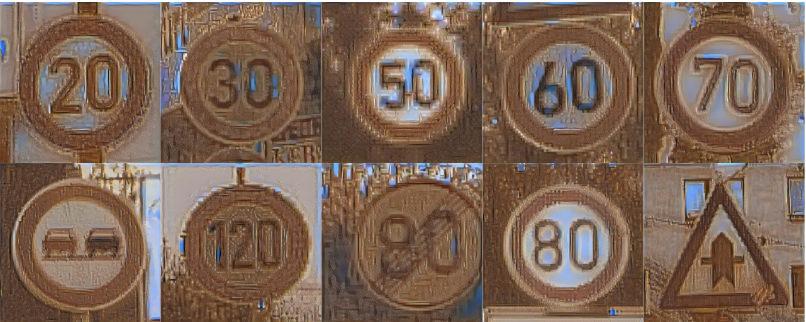}
    \caption{Examples of ST-Augmented Images}
    \label{fig:augmented_images}
\end{figure}

\subsection{Results}

\begin{figure}
    \centering
    \begin{subfigure}{0.8\linewidth}
        \includegraphics[width=\linewidth]{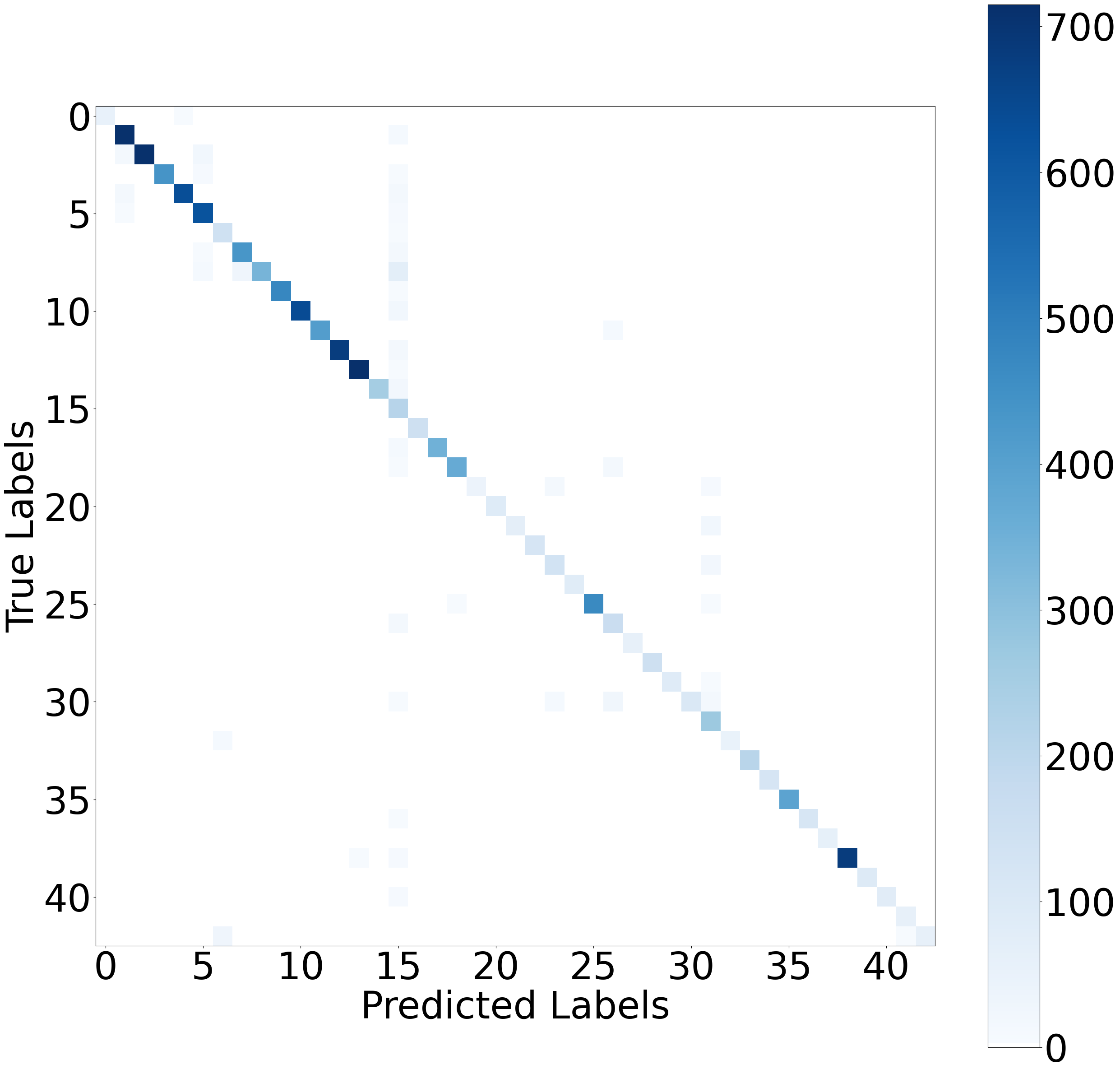}
        \caption{Experiment 3}
        \label{fig:sub1}
    \end{subfigure}
    \\
    \begin{subfigure}{0.8\linewidth}
        \includegraphics[width=\linewidth]{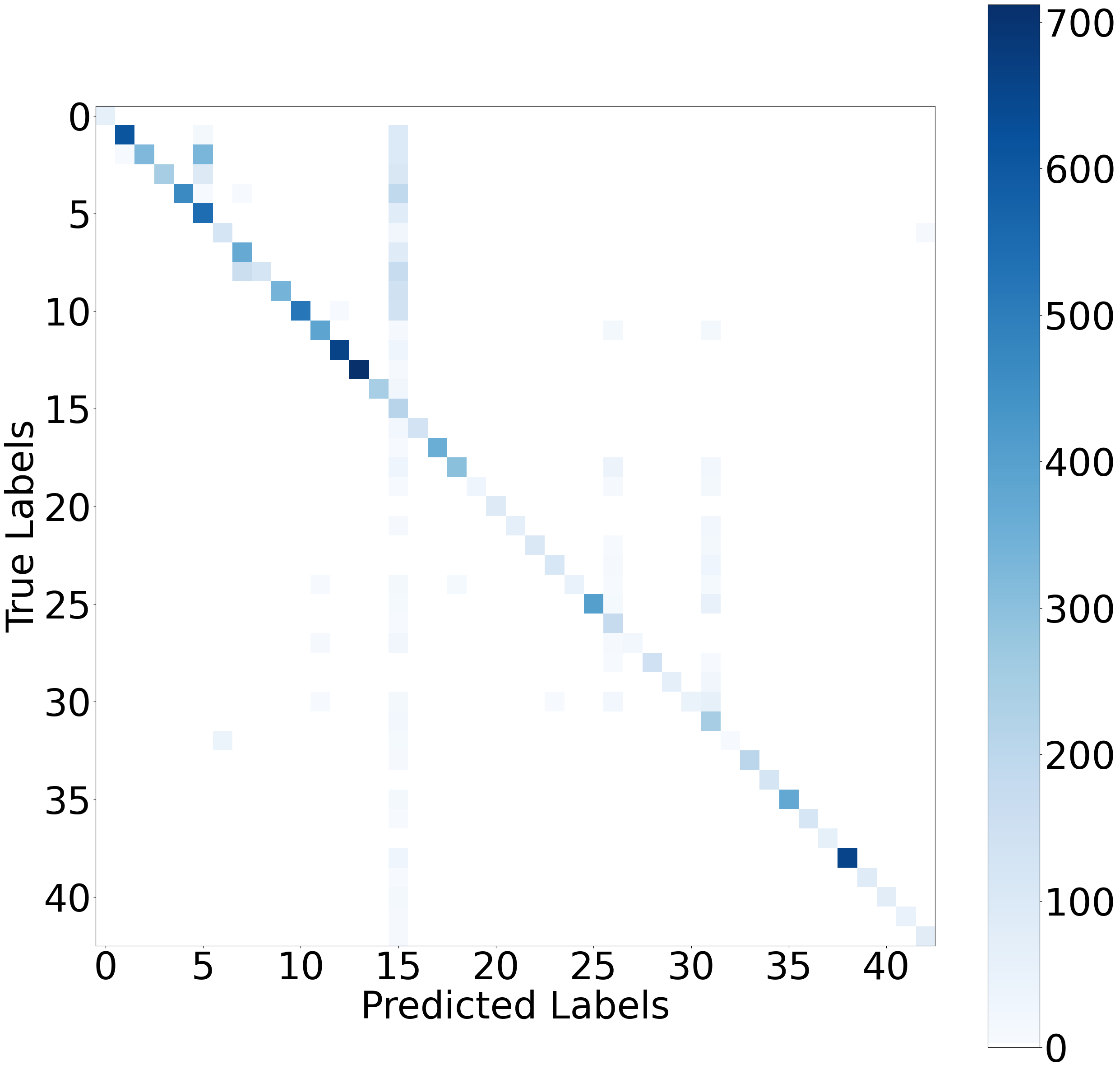}
        \caption{Experiment 4}
        \label{fig:sub2}
    \end{subfigure}
    \caption{Confusion matrices for the model retrained using augmented datasets (Orig + ST-Aug)}
    \label{fig:conf_matrices}
\end{figure}

We compare models trained only on the standard GTSRB training set versus using our proposed runtime adaptation approach. Figure ~\ref{fig:augmented_images} shows sample instances of ST-augmented prototypes. Adapting the model by training it on ST-augmented instances extracted from the degraded test set significantly improves robustness, improving the model from Experiment 2 (F1=0.63) to Experiment 4 (F1=0.86). Table \ref{tbl:summary_performance} summarizes the system's performance metrics across different experiments. Figure \ref{fig:conf_matrices} visualize the confusion matrices that highlight heavy misclassification in a few classes. 

\begin{enumerate}
    \item Experiment 1: The model shows the best performance under familiar conditions with an F1 score of 0.97 and an accuracy of 0.96.
    
    \item Experiment 2: The model trained on original images struggles when tested on augmented data, reflecting a notably diminished recall of 0.56 and a corresponding F1 score of 0.63. This accentuates the model's challenges when confronted with unfamiliar data.
    
    \item Experiment 3: By training on a combination of both original and augmented images and testing on original images, the model maintains a commendable performance, evidenced by an F1 score of 0.93. This implies that integrating augmented data into the training doesn't compromise the model's performance in identifying original images.
    
    \item Experiment 4: With the same mixed training set, but subjected to testing on augmented images, the model manifests resilience. The F1 score stays relatively elevated at 0.86, underscoring the model's adaptability to previously unseen or infrequent data variations.
    
    \item Experiment 5: When exclusively trained on augmented images and tested on their original counterparts, there's a noticeable decline in the model's performance, with an F1 score resting at 0.58. This outcome emphasizes the importance of incremental augmentation.
    
    \item Experiment 6: A parallel trend is observable when evaluating augmented data. Nevertheless, the performance here is more satisfactory (F1 score of 0.75) than in Experiment 5. This reaffirms the inherent distinctiveness of features in augmented images.
\end{enumerate}

\begin{table}
	\caption{The performance summary for all six experiments}\label{tbl:summary_performance}
	\begin{center}
		\begin{tabular}{|c| l | l | c | c | c | c|  }
                        \hline
			Exp. &  Train           & Test & Precision      & Recall & F1 & Accuracy\\ 
                        \hline
			1 &  Orig                & Orig & 0.98            & 0.96   & 0.97 & 0.96 \\
			2 &  Orig                & ST-Aug & 0.89            & 0.56   & \textbf{0.63} & 0.56\\
			3 &  Orig + ST-Aug       & Orig & 0.95            & 0.92   & 0.93 & 0.92 \\
			4 &  Orig + ST-Aug       & ST-Aug & 0.92            & 0.84   & \textbf{0.86} & 0.85 \\
			5 &  ST-Aug              & Orig & 0.91            & 0.52   & 0.58 & 0.52\\
			6 &  ST-Aug              & ST-Aug & 0.91            & 0.69   & 0.75 & 0.70\\
		        \hline
		\end{tabular}
	\end{center}
\end{table}

\section{Discussion}

\subsection{Improving Model Robustness with Trade-offs}

In Experiment 5, the model trained solely on rusty images (augmented data) performed poorly when tested on original images, yielding an F1 score of 0.58. This indicates that the features of rusty images are quite distinct from those of the original images. In Experiment 2, the model trained only on original images produced a subpar F1 score of 0.63 when tested on augmented images. This reaffirms that the original training data is inadequate to effectively capture the features of the rusty images. Together, these results from Experiments 5 and 2 highlight the substantial feature differences between original and rusty images, validating the idea that rusty images can indeed be treated as unknown-unknown scenarios in real-world applications.

Experiment 1 confirms that the model performs exceptionally well in familiar conditions. However, the poorer performance of Experiment 2 indicates that models trained solely on original images struggle when confronted with augmented data. This highlights the model's inability to generalize effectively to unseen or rare variations in data. In other words, without exposure to the augmented scenarios during training, the model is ill-equipped to handle real-world anomalies and unknown-unknown situations. 

By observing the results from Experiment 3 (F1=0.93) and Experiment 1 (F1=0.97), it is evident that introducing rusty images into the training set does not hinder the model's performance on original test images. Looking at Experiment 4, despite a slight decrease in the F1 score (F1=0.86) compared to Experiment 3, this is still relatively high compared to Experiment 2 (F1=0.63). Training on both original and augmented images but testing on augmented images demonstrates that the model is robust, even though there's some room for improvement.

The positive effects of data augmentation using neural style transfer are evident. While there's a slight trade-off in accuracy on original images, the model exhibits enhanced adaptability to the novel situation, generated rusty images. The marginal 4\% reduction in F1 score between Experiment 1 and Experiment 3 along with the significant 23\% improvement in F1 scores from Experiment 2 to Experiment 4, reaffirm the effectiveness of our method in enhancing robustness against unknown-unknown scenarios.

\subsection{Incremental Data Augmentation}

Incremental data augmentation, where original data is supplemented with style transfer-augmented images, has emerged as a critical strategy to improve model robustness. Our experimental results solidify this perspective. When models were trained exclusively on style transfer-augmented (ST-Augmented) images, as depicted in Experiment 5, their performance suffered, yielding an F1 score of 0.58 when assessed against original images. Moreover, when we evaluated the performance of models trained only on ST-Augmented data and tested on augmented images, as seen in Experiment 6, the F1 score was a middling 0.75. This contrast not only stresses the potential pitfalls of relying solely on augmented data but also accentuates the salient benefits of integrating original images during training. Incremental data augmentation ensures the model remains versatile and effective across a diverse range of real-world scenarios.

\subsection{Future Improvements}

\begin{figure}
    \centering
    \includegraphics[width=0.8\linewidth]{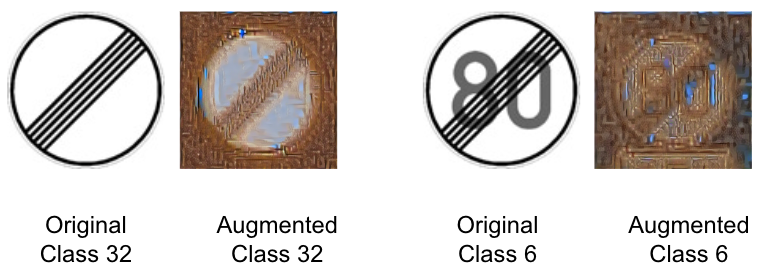}
    \caption{Examples of ST-Augmented Images}
    \label{fig:confused_class}
\end{figure}

The application of NST to generate augmented (rusty) images brings about both promising results and areas of improvement. While NST has demonstrated success across most classes, we observed challenges in a few specific classes. There exists a risk that NST, in its pursuit of augmentation, excessively transforms the original images, leading to the production of augmented data that may no longer retain salient features of the original data. Such over-modification could result in what we term 'uninformative' augmented images. This results in the model's predisposition to confuse signs. For example, the heatmap of confusion matrices shown in Fig.~\ref{fig:conf_matrices} indicates confusion between class 6 and class 32, even though Fig.~\ref{fig:confused_class} shows that both classes are originally distinctive.

The "Catch-All Phenomenon", e.g., exhibited by class 15 which is highlighted as vertical colored blocks on the 15th column of Fig.~\ref{fig:conf_matrices}, underscores a pivotal challenge in our data augmentation approach. This indicates that while the model accurately identifies the majority of classes, many augmented images become 'uninformative' post-augmentation. Consequently, these 'uninformative' instances are frequently misclassified, often being merged into broader and more generic categories, such as class 15. When a significant portion of these augmented instances are incorrectly pooled into catch-all categories, many true instances of other classes are not correctly identified, leading to a diminished overall recall across the dataset. 

Such inconsistencies in ST-augmentation arise due to the nuances in the original images, e.g. class 32, which may not always be optimally augmented using the current NST methodology. Another contributing factor could be the non-discriminative approach where not only is the sign subjected to the augmentation process, but the background is also altered. Such an all-encompassing transformation can lead to inaccuracies in the generated images. 

Moving forward, there is a clear need to fine-tune NST parameters. Employing pre-processing steps, such as image segmentation, could ensure the isolation and preservation of pivotal features. Additionally, post-processing might further refine the augmented output, rectifying any minor discrepancies induced by NST. This will ensure critical features remain intact and are not lost during the transformation process. The potential of this approach goes beyond rusty images. The framework can be expanded to cater to a myriad of real-world degradations such as graffiti on signs, physical damage, and other common degradations. Addressing these will not only improve the performance of the model but also push the boundaries of robust autonomy in unstructured environments. 

\section{Conclusion}

In our study, we explored the implications of data augmentation using neural style transfer on model performance. Although the neural style transfer was effective for most classes, specific classes need tailored approaches. The introduction of generated rusty images to the training set showed enhanced robustness against unknown-unknown scenarios, with only a minor trade-off in accuracy on original images. 

While the outcomes are promising, the challenges posed by uninformative augmented images and model confusion among similar signs accentuate the need for further research in advanced augmentation techniques and more discriminative model training. This work paves the way for subsequent studies targeting a comprehensive understanding and better handling of real-world data aberrations.

\bibliographystyle{IEEEtran}
\bibliography{references}

\end{document}